\documentclass[10pt,twocolumn,letterpaper]{article}

\usepackage[pagenumbers]{cvpr} 

\definecolor{cvprblue}{rgb}{0.21,0.49,0.74}
\usepackage[pagebackref,breaklinks,colorlinks,allcolors=cvprblue]{hyperref}

\usepackage{graphicx}
\usepackage{amsmath}
\usepackage{multirow}
\usepackage{cuted}
\usepackage[misc]{ifsym}

\definecolor{blue}{HTML}{0055cc}
\definecolor{red}{HTML}{cc1100}
\definecolor{orange}{HTML}{cc7700}
\definecolor{gray}{HTML}{efefef}
\definecolor{darkgreen}{rgb}{0.13, 0.55, 0.13}
\definecolor{darkgray}{HTML}{757575}

\newcommand{\figref}[1]{Figure~\ref{#1}}
\newcommand{\tabref}[1]{Table~\ref{#1}}
\newcommand{\secref}[1]{Section~\ref{#1}}

\title{SAMPart3D: Segment Any Part in 3D Objects}

\author{Yunhan Yang$^{1}$
\qquad
Yukun Huang$^{1}$
\qquad
Yuan-Chen Guo$^{2}$
\qquad
Liangjun Lu$^{1}$
\\
Xiaoyang Wu$^{1}$
\qquad
Edmund Y. Lam$^{1}$
\qquad
Yan-Pei Cao$^{2\dag}$
\qquad
Xihui Liu$^{1}\textsuperscript{\Letter}$
\vspace{0.2cm}
\\
{\normalsize $^{1}$ The University of Hong Kong \quad $^{2}$ VAST} \\
{\normalsize Project Page: {\href{https://yhyang-myron.github.io/SAMPart3D-website}{https://yhyang-myron.github.io/SAMPart3D-website}}}
}

\begin{document}
\maketitle

\let\thefootnote\relax\footnotetext{\Letter~: Corresponding author, \dag: Project leader.}

\begin{strip}
    \centering
    \vspace{-4em}
    \includegraphics[width=\textwidth]{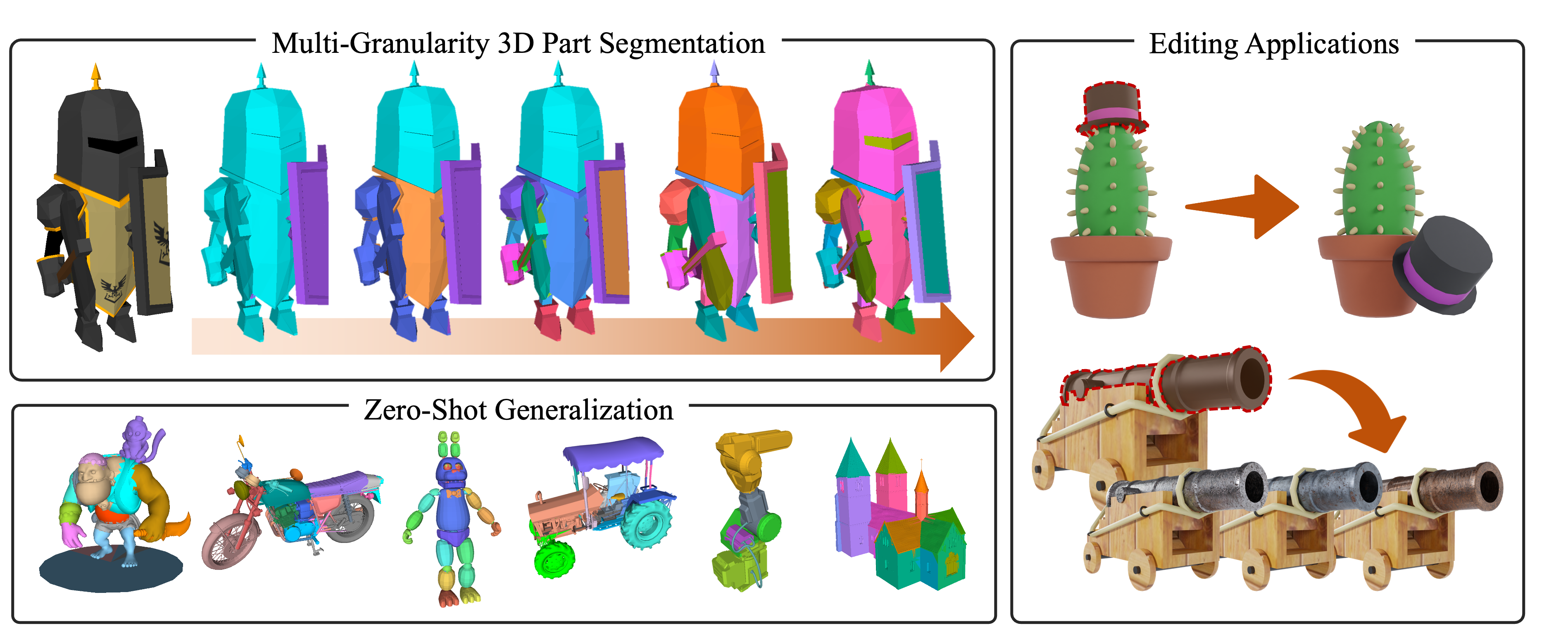}
    \captionof{figure}{\textbf{SAMPart3D} is able to segment any 3D object into semantic parts across multiple levels of granularity, without the need for predefined part label sets or text prompts. It supports a range of applications, including part-level editing and interactive segmentation.}
    \label{fig:teaser}
\end{strip}

\begin{abstract}
3D part segmentation is a crucial and challenging task in 3D perception, playing a vital role in applications such as robotics, 3D generation, and 3D editing. Recent methods harness the powerful Vision Language Models (VLMs) for 2D-to-3D knowledge distillation, achieving zero-shot 3D part segmentation.
However, these methods are limited by their reliance on text prompts, which restricts the scalability to large-scale unlabeled datasets and the flexibility in handling part ambiguities. In this work, we introduce \textbf{SAMPart3D}, a scalable zero-shot 3D part segmentation framework that segments any 3D object into semantic parts at multiple granularities, without requiring predefined part label sets as text prompts. For scalability, we use text-agnostic vision foundation models to distill a 3D feature extraction backbone, allowing scaling to large unlabeled 3D datasets to learn rich 3D priors. For flexibility, we distill scale-conditioned part-aware 3D features for 3D part segmentation at multiple granularities. Once the segmented parts are obtained from the scale-conditioned part-aware 3D features, we use VLMs to assign semantic labels to each part based on the multi-view renderings. Compared to previous methods, our SAMPart3D can scale to the recent large-scale 3D object dataset Objaverse and handle complex, non-ordinary objects. Additionally, we contribute a new 3D part segmentation benchmark to address the lack of diversity and complexity of objects and parts in existing benchmarks. Experiments show that our SAMPart3D significantly outperforms existing zero-shot 3D part segmentation methods, and can facilitate various applications such as part-level editing and interactive segmentation.
\end{abstract}

\vspace{-1em}
\section{Introduction}

3D part segmentation is a fundamental 3D perception task that is essential for various application areas, such as robotic manipulation, 3D analysis and generation, part-level editing~\cite{tertikas2023generating} and stylization~\cite{chung20243dstyleglip}.

In the past few years, data-driven fully supervised methods~\cite{qi2017pointnet, qi2017pointnet++, qian2022pointnext, zhao2021point, li2018pointcnn} have achieved excellent results on closed-set 3D part segmentation benchmarks~\cite{mo2019partnet, chang2015shapenet}. However, these methods are limited to segmenting simple objects due to the restricted quantity and diversity of 3D data with part annotations. Despite the recent release of large-scale 3D object datasets~\cite{deitke2023objaverse, deitke2024objaverse, wu2023omniobject3d}, acquiring part annotations for such vast amounts of 3D assets is time-consuming and labor-intensive, which prevents 3D part segmentation from replicating the success of data scaling and model scaling in 2D segmentation~\cite{kirillov2023segment}.

To achieve zero-shot 3D part segmentation in the absence of annotated 3D data, several challenges need to be addressed. The first and most significant challenge is \textbf{\emph{how to generalize to open-world 3D objects without 3D part annotations}}. To tackle this, recent works~\cite{liu2023partslip,zhou2023partslip++,kim2024partstad,abdelreheem2023satr,xue2023zerops} have utilized pre-trained 2D foundation vision models, such as SAM~\cite{kirillov2023segment} and GLIP~\cite{li2022glip}, to extract visual information from multi-view renderings and project it onto 3D primitives, achieving zero-shot 3D part segmentation. However, these methods rely solely on 2D appearance features without 3D geometric cues, leading to the second challenge: \textbf{\emph{how to leverage 3D priors from unlabeled 3D shapes}}. PartDistill~\cite{umam2023partdistill} has made a preliminary exploration by introducing a 2D-to-3D distillation framework to learn 3D point cloud feature extraction, but it cannot scale to large 3D datasets like Objaverse~\cite{deitke2023objaverse} due to the need for predefined part labels and the constrained capabilities of GLIP. Building on existing works, we further explore the third challenge: \textbf{\emph{the ambiguity of 3D parts}}, which manifests primarily in semantics and granularity. \emph{Semantic ambiguity} arises from the vague textual descriptions of parts. Existing methods rely on vision-language models (VLMs) like GLIP, which require a part label set as text prompt. Unfortunately, not all 3D parts can be clearly and precisely described in text. \emph{Granularity ambiguity} considers that a 3D object can be segmented at multiple levels of granularity. For example, the human body can be divided into broader sections, such as upper and lower halves, or into finer parts like limbs, torso, and head. Previous methods rely on fixed part label sets and lack flexible control over segmentation granularity.

To tackle the three aforementioned challenges, in this work, we propose \textbf{SAMPart3D}, a scalable zero-shot 3D part segmentation framework that segments object parts at multiple granularities without requiring preset part labels as text prompts.
We argue that previous works overly rely on predefined part label sets and GLIP, limiting their scalability to complex, unlabeled 3D datasets and their flexibility in handling semantic ambiguity of 3D parts. To address this, we abandon GLIP and instead utilize the more low-level, text-independent DINOv2~\cite{oquab2023dinov2} model for 2D-to-3D feature distillation, eliminating the reliance on part label sets and enhancing both scalability and flexibility. Besides, to handle the ambiguity in segmentation granularity, we employ a scale-conditioned MLP~\cite{kim2024garfield} distilled from SAM for granularity-controllable 3D part segmentation. The distillation from DINOv2 and SAM is divided into two training stages to balance efficiency and performance. After obtaining the segmented 3D parts, we adaptively render multi-view images for each part based on its visual area, then use the powerful Multi-modal Large Language Models (MLLMs)~\cite{liu2024visual, chen2023internvl, wang2023cogvlm, he2024bunny} to assign semantic descriptions for each part based on the renderings, yielding the final part segmentation results.

In summary, our contributions are as follows: 
\begin{itemize}
\item We introduce SAMPart3D, a scalable zero-shot 3D part segmentation framework that segments object parts at multiple granularities without requiring preset part labels as text prompts.
\item We propose a text-independent 2D-to-3D distillation, which enables learning 3D priors from large-scale unlabeled 3D objects and can handle part ambiguity in both semantic and granularity aspects. The distillation is two-stage, striking a balance between segmentation performance and training efficiency.
\item We introduce PartObjaverse-Tiny, a 3D part segmentation dataset which provides detailed semantic and instance level part annotations for 200 complex 3D objects.
\item Extensive experiments demonstrate that SAMPart3D achieves outstanding part segmentation results on complex and diverse 3D objects compared to existing zero-shot 3D part segmentation methods. Furthermore, our method can facilitate various applications, such as interactive segmentation and part-level editing.
\end{itemize}

\section{Related Work}

\begin{figure*}
\centering
\includegraphics[width=\linewidth]{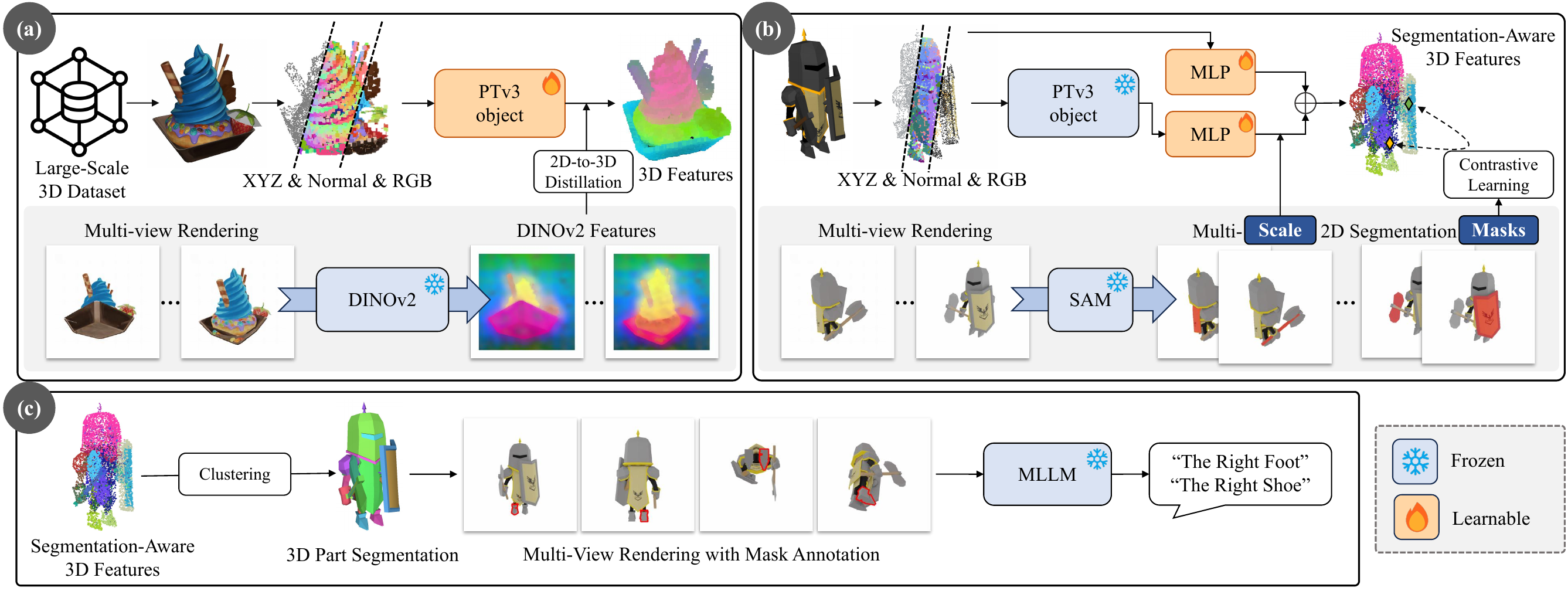}
\caption{An overview pipeline of \textbf{SAMPart3D}. (a) We first pre-train 3D backbone PTv3-object on 3D large-scale data Objaverse, distilling visual features from FeatUp-DINOv2. (b) Next, we train light-weight MLPs to distill 2D masks to scale-conditioned grouping. (c) Finally, we cluster the feature of point clouds and highlight the consistent 2D part area with 2D-3D mapping on multi-view renderings, and then query semantics from MLLMs.}
\vspace{-13px}
\label{fig:pipeline}
\end{figure*}

\noindent\textbf{2D Foundation Models.} 
Recently, 2D vision foundation models have advanced significantly due to large-scale data and model size growth. Based on learning strategies, these models can be grouped into: \emph{traditional models}, \emph{textually-prompted models}, and \emph{visually-prompted models}. \emph{Traditional models} rely solely on images and use self-supervised objectives, like masked patch reconstruction in MAE~\cite{he2022masked} and self-distillation in DINO~\cite{caron2021dino,oquab2023dinov2}. \emph{Textually-prompted models} use large-scale text-image pairs, as seen in CLIP~\cite{radford2021clip}, which aligns images and text for strong zero-shot performance. Since text prompts are less effective for fine-grained tasks like segmentation~\cite{wei2024ov}, \emph{visually-prompted models} use visual cues like bounding boxes, points, or masks. SAM~\cite{kirillov2023segment}, for example, employs visual prompts for zero-shot segmentation on new domains. Recently, efforts~\cite{zhang2022pointclip, zhu2023pointclipv2, huang2023clip2point, chen2023clip2scene, yang2023sam3d, yin2024sai3d, qi2024gpt4point} have explored using 2D foundation models for 3D content understanding. In this work, we integrate multiple 2D vision foundation models for zero-shot 3D semantic part segmentation at varying granularities.

\noindent\textbf{3D Part Segmentation.} 3D part segmentation aims to divide a 3D object into semantic parts, which is a long-standing problem in 3D computer vision. Early works~\cite{qi2017pointnet,qi2017pointnet++,qian2022pointnext,zhao2021point,li2018pointcnn} primarily focused on exploring network architecture designs to better learn 3D representations. Qi et al.~\cite{qi2017pointnet++} propose a hierarchical neural network named PointNet++, which leverages neighborhoods at multiple scales to achieve both robustness and detail capture. Zhao et al.~\cite{zhao2021point} design an expressive Point Transformer layer, which can be used to construct high-performing backbones for semantic segmentation of 3D point clouds. These methods typically employ fully supervised training~\cite{zhao2021point}, requiring time-consuming and labor-intensive 3D part annotations. Limited by the scale and diversity of 3D part datasets~\cite{mo2019partnet,chang2015shapenet}, they struggle to achieve generalization on complex 3D objects in open-world scenarios.

\begin{figure*}
\centering
\includegraphics[width=\linewidth]{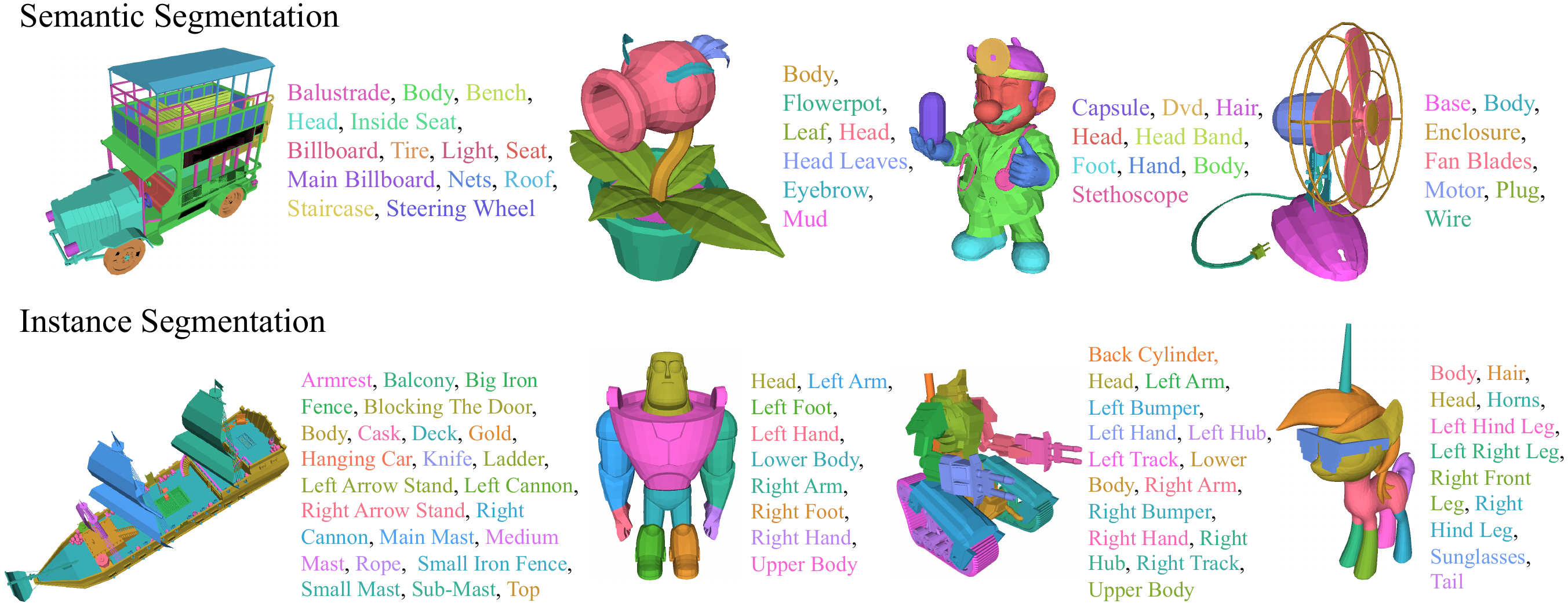}
\caption{Visualization of \textbf{PartObjaverse-Tiny} with part-level semantic and instance segmentation labels.}
\vspace{-1em}
\label{fig:partobj_vis}
\end{figure*}

\noindent\textbf{Zero-shot 3D Part Segmentation.} To overcome the the limitations of 3D annotated data and pursue zero-shot capabilities, recent 3D part segmentation methods~\cite{liu2023partslip,zhou2023partslip++,umam2023partdistill,xue2023zerops,abdelreheem2023satr,zhu2023pointclipv2,tang2024segment,thai20243x2,zhong2024meshsegmenter} leverage 2D priors from foundation vision models~\cite{oquab2023dinov2,radford2021clip,li2022glip,kirillov2023segment}.
PartSLIP~\cite{liu2023partslip} leverages the image-language model GLIP~\cite{li2022glip} to solve both semantic and instance segmentation for 3D object parts, where the GLIP model is expected to predict multiple bounding boxes for all part instances. The subsequent PartSLIP++~\cite{zhou2023partslip++} integrates the pretrained 2D segmentation model SAM~\cite{kirillov2023segment} into the PartSLIP pipeline, yielding more accurate pixel-wise part annotations than the bounding boxes used in PartSLIP.
Similarly, ZeroPS~\cite{xue2023zerops} introduces a two-stage 3D part segmentation and classification pipeline, bridging the multi-view correspondences and the prompt mechanism of foundational models.
Unlike previous methods directly transferring 2D pixel-wise or bounding-box-wise predictions to 3D segmentation, PartDistill~\cite{umam2023partdistill} adopts a cross-modal teacher-student distillation framework to learn a 3D student backbone for extracting point-specific geometric features from unlabeled 3D shapes.
In this work, we extend cross-modal distillation to a more challenging large-scale 3D dataset, Objaverse~\cite{deitke2023objaverse}, without requiring text prompts for 3D parts, and achieve granularity-controllable segmentation.

\section{Method}

As shown in \figref{fig:pipeline}, the proposed framework \textbf{SAMPart3D} consists of three stages: \textbf{\emph{large-scale pre-training}} to learn a 3D feature extraction backbone from a vast number of unlabeled 3D objects, as described in \secref{sec:method-pretrain}; \textbf{\emph{sample-specific fine-tuning}} to train a lightweight MLP for scale-conditioned grouping, as described in \secref{sec:method-grouping_field}; and training-free \textbf{\emph{semantic querying}} for assigning semantic labels to each part using a multimodal large language model, as described in \secref{sec:method-semantic}.

\subsection{Large-scale Pre-training: Distilling 2D Visual Features to 3D Backbone}
\label{sec:method-pretrain}

In this stage, we aim to learn a 3D feature extraction backbone that leverages the geometric cues of 3D objects and learns 3D priors from a large-scale collection of unlabeled 3D objects.

\noindent\textbf{Training Data.} Unlike the previous work PartDistill~\cite{umam2023partdistill}, which was trained on limited categories of 3D objects, we utilize the large-scale 3D object dataset Objaverse~\cite{deitke2023objaverse} as our training data. Objaverse encompasses over 800K 3D assets spanning diverse object categories, providing rich 3D priors for zero-shot 3D part segmentation. Additionally, considering that current 3D feature extraction network architectures are primarily developed for 3D point clouds, we randomly sample point clouds from the mesh surfaces of 3D objects as the input to our backbone.

\noindent\textbf{Backbone for 3D Feature Extraction.}
Building upon the state-of-the-art point cloud perception backbone, Point Transformer V3 (PTv3)~\cite{wu2023point, pointcept2023}, we further tailor the architecture to accommodate the characteristics of 3D objects, resulting in PTv3-object. Specifically, PTv3 is designed for scene-level point clouds, incorporating numerous down-sampling layers for a large receptive field and low computational load. However, the number of points and spatial extent required to represent an object is much smaller than that required for a scene. Therefore, we removed most of the down-sampling layers from PTv3 and instead stacked more transformer blocks to enhance detail preservation and feature abstraction. Note that our learning framework is model-agnostic and can be used with other more advanced network architectures for 3D feature extraction.

\noindent\textbf{Distilling 2D Visual Features to 3D Backbone.}
To train the backbone for 3D feature extraction on large-scale unlabeled 3D objects from Objaverse~\cite{deitke2023objaverse}, pre-trained 2D vision foundation models are needed as supervision. Previous methods, such as PartDistill~\cite{umam2023partdistill}, use VLMs as supervision and require part label sets as text prompts, making it difficult to scale to Objaverse. Therefore, we abandon VLMs and instead utilize the more low-level, text-independent DINOv2~\cite{oquab2023dinov2} model as supervision for visual feature distillation. In particular, the visual features extracted by DINOv2 are low-resolution and lack detail, making them unsuitable for subsequent part segmentation. To address this, we employ the recently proposed feature upsampling technique, FeatUp~\cite{fu2024featup}, to enhance the DINOv2 features for use as point-wise supervision in 3D feature extraction.

Specifically, for each training iteration, we sample a batch of 3D objects, with each object represented by a point cloud $ X \in \mathbb{R}^{N \times 3} $, where $N$ denotes the number of 3D points. We input the point cloud $X$ into the PTv3-object backbone, resulting in 3D features $ F_\text{3D} \in \mathbb{R}^{N \times C} $, where $C$ is the feature dimension of $384$ consistent with DINOv2 features. Then, to obtain corresponding 2D visual features for supervision, we render  images from $K$ different views for each object and extract the corresponding DINOv2 features. Utilizing the mapping relationship between point clouds and pixels, we can directly obtain the 2D features $ F_\text{2D} \in \mathbb{R}^{N \times C} $ of the 3D point cloud. However, considering occlusion, not all 3D points can be assigned 2D features given a single view. To address this, we use depth information to determine the occlusion status of the point cloud following~\cite{hu2021bidirectional}. For occluded 3D points, we directly assign their original 3D features from $ F_\text{3D} $.
Finally, by averaging the 2D features from all $K$ rendered views, we obtain the final 2D features of the point cloud:
\begin{equation}
F_\text{2D} = \frac{1}{K} \sum_{k=1}^{K} F_\text{2D}^{(k)},
\end{equation}
where $F_\text{2D}^{(k)}$ represents the obtained 2D features of the point cloud at the $k$-th view, and we simply choose a mean squared error (MSE) loss:
\begin{equation}
\mathcal{L_\text{pre}} = (F_\text{3D} - F_\text{2D})^2
\end{equation}
as the learning objective for distilling 2D visual features to the 3D backbone.

\subsection{Sample-specific Fine-tuning: Distilling 2D Masks for Multi-granularity Segmentation}
\label{sec:method-grouping_field}

After pre-training the backbone via distilling the 2D visual features, we can effectively extract 3D features of any 3D object. These 3D features are used together with 2D segmentation masks from SAM~\cite{kirillov2023segment} for zero-shot 3D part segmentation. Furthermore, considering the ambiguity in segmentation granularity, we aim to introduce a \textit{scale} value to control the granularity of the segmentation. To this end, we introduce a scale-conditioned lightweight MLP that enables 3D part segmentation at various scales, inspired by GARField~\cite{kim2024garfield} and GraCo~\cite{zhao2024graco}.

\noindent\textbf{Long Skip Connection.}
Although the pre-trained backbone is able to extract rich 3D features, the low-level cues of point cloud (critical for point-wise prediction tasks) are lost due to overly deep networks. Therefore, we introduce a MLP-based long skip connection module to capture the low-level features of point cloud. Specifically, we first assign the normal values of the faces to each corresponding point in the point cloud to provide the shape information of the mesh. Then, these normal values, along with color and coordinates, serve as inputs for the long skip connection module, the outputs of which are added to the outputs of the 3D backbone to complement low-level features.

\noindent\textbf{Scale-conditioned Grouping.} We first render multi-view images of the 3D object and utilize SAM to generate 2D masks of these multi-view renderings. For each mask, we can find the relevant points and calculate the 3D scale $\sigma$ with:
\begin{equation}
\sigma = \sqrt{(\varepsilon \sigma_x)^{2} + (\varepsilon \sigma_y)^{2} + (\varepsilon \sigma_z)^{2}} ,
\end{equation}
where $\sigma_x, \sigma_y, \sigma_z$ are the standard deviations of coordinates in the $x, y, z$ directions, respectively; $\varepsilon$ is a scaling factor for better distinguishing the scales of different masks, which we set to 10. 

Then, we sample paired pixels on the valid region of 2D renderings for contrastive learning. Specifically, for two 3D points $p_i$ and $p_j$ mapping from a 2D pixel pair, we can obtain their features:
\begin{equation}
\mathbf{F}_i = \mathbf{F}_i^B(\sigma_i) + \mathbf{F}_i^P(\sigma_i) ,~~~~ \mathbf{F}_j = \mathbf{F}_j^B(\sigma_j) + \mathbf{F}_j^P(\sigma_j) ,
\end{equation}
where $\mathbf{F}^B(\sigma)$ is the feature derived from backbone PTv3-object, and $\mathbf{F}^P(\sigma)$ represents the positional embedding derived from positional encoding module. The final contrastive loss is:
\begin{equation}
\mathcal{L}_{con} = \begin{cases} 
\left \|\mathbf{F}_i - \mathbf{F}_j\right \| , & \text{if } \mathcal{C}(i,j) = 1 \\
\text{ReLU}(m - \left \|\mathbf{F}_i - \mathbf{F}_j\right \|), & \text{if } \mathcal{C}(i,j) = 0
\end{cases}
\end{equation}
where $\mathcal{C}(i,j)$ is a binary function that indicates whether the pair $(i,j)$ is from the same mask (1) or different masks (0), and $m$ is a lower margin.

After training the scale-conditioned MLP, we can obtain the segmentation-aware features of 3D point cloud conditioned on a scale. By applying clustering algorithms such as HDBSCAN~\cite{mcinnes2017hdbscan} to these grouping features, we can segment the 3D point cloud into different parts. The segmentation of 3D mesh can be easily derived from the segmentation of 3D point cloud using a voting algorithm.

\subsection{Semantic Querying with MLLMs}
\label{sec:method-semantic}

After obtaining the part segmentation results of a 3D object, we query the semantics label of each part using the powerful Multimodal Large Langauge Models (MLLMs), as shown in \figref{fig:pipeline} (c). Utilizing the 3D-to-2D mapping, we can identify the corresponding 2D area of each 3D part in the multi-view renderings, which enables view-consistent highlighting of 3D parts in 2D renderings. By inputting these highlighted results into MLLMs, we can perform per-part semantic querying.

Specifically, we first select several canonical views of an object for rendering and part highlighting. This enriches the object details in the rendered images and facilitates the perception of MLLMs. Next, we choose a view with the largest rendered area for the part of interest and highlight the corresponding part area, ensuring comprehensive incorporation of this part’s details. Finally, we combine these images and feed them into MLLMs to obtain the semantic labels of the part.

\begin{figure*}
\centering
\includegraphics[width=\linewidth]{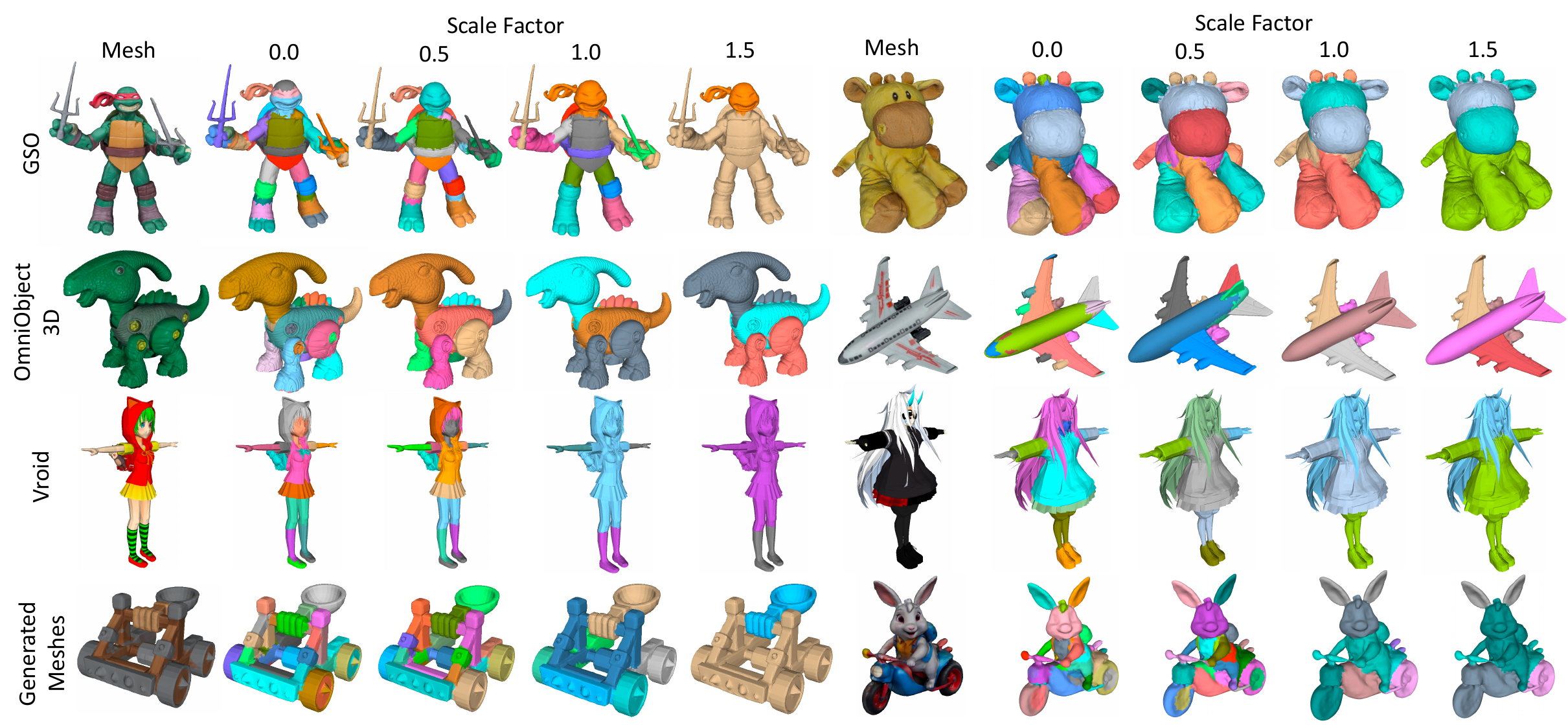}
\caption{Visualization of multi-granularity 3D part segmentation on GSO~\cite{downs2022google}, OmniObject3D~\cite{wu2023omniobject3d}, Vroid~\cite{chen2023panic3d} and 3D generated meshes.}
\label{fig:exp_scale}
\vspace{-10pt}
\end{figure*}
\section{Experiments}

\subsection{PartObjaverse-Tiny}
Current part segmentation datasets typically include limited object categories and incomplete part annotations. 
This makes existing datasets unsuitable for evaluating the 3D segmentation performance of arbitrary objects and parts.
Therefore, we annotate a subset of Objaverse, named \textbf{\emph{PartObjaverse-Tiny}}, which consists of 200 shapes with fine-grained annotations. Following GObjaverse~\cite{qiu2023richdreamer}, we divide these 200 objects into 8 categories: Human-Shape (29), Animals (23), Daily-Used (25), Buildings\&\&Outdoor (25), Transportations (38), Plants (18), Food (8) and Electronics (34). Each major category includes multiple smaller object categories, for example, Transportation includes cars, motorcycles, airplanes, cannons, ships, among others. For each object, we meticulously segment and annotate it into fine-grained, semantically coherent parts. We present examples of PartObjaverse-Tiny dataset for semantic segmentation and instance segmentation in \figref{fig:partobj_vis}.

\subsection{Experiments Results}
\noindent\textbf{Multi-granularity 3D part segmentation.} To demonstrate the generalization ability of our model, we use the model pretrained on Objaverse~\cite{deitke2023objaverse} to segment objects in GSO~\cite{downs2022google}, OmniObject3D~\cite{wu2023omniobject3d} and Vroid~\cite{chen2023panic3d} datasets, as well as on 3D meshes generated from TripoAI~\cite{TripoAI-website} and Rodin~\cite{Rodin-website}. Multi-granularity segmentation results are shown in \figref{fig:exp_scale}.

\noindent\textbf{Metrics.} We utilize class-agnostic mean Intersection over Union (mIoU) to evaluate part segmentation results without semantics. Follow \cite{wang2021learning, xue2023zerops}, for each ground-truth part, we calculate the IoU with every predicted part, and assign the maximum IoU as the part IoU. Finally, we calculate the average of part IoU as class-agnostic mIoU. For semantic evaluation, we follow \cite{mo2019partnet,liu2023partslip}, utilizing category mIoU and mean Average Precision (mAP) with a 50\% IoU threshold as metrics for semantic and instance segmentation, respectively. We consider each object as a separate category, calculate the IoU/AP50 of each part separately, and compute the mIoU/mAP50 of this object.

\begin{figure*}
\centering
\includegraphics[width=\linewidth]{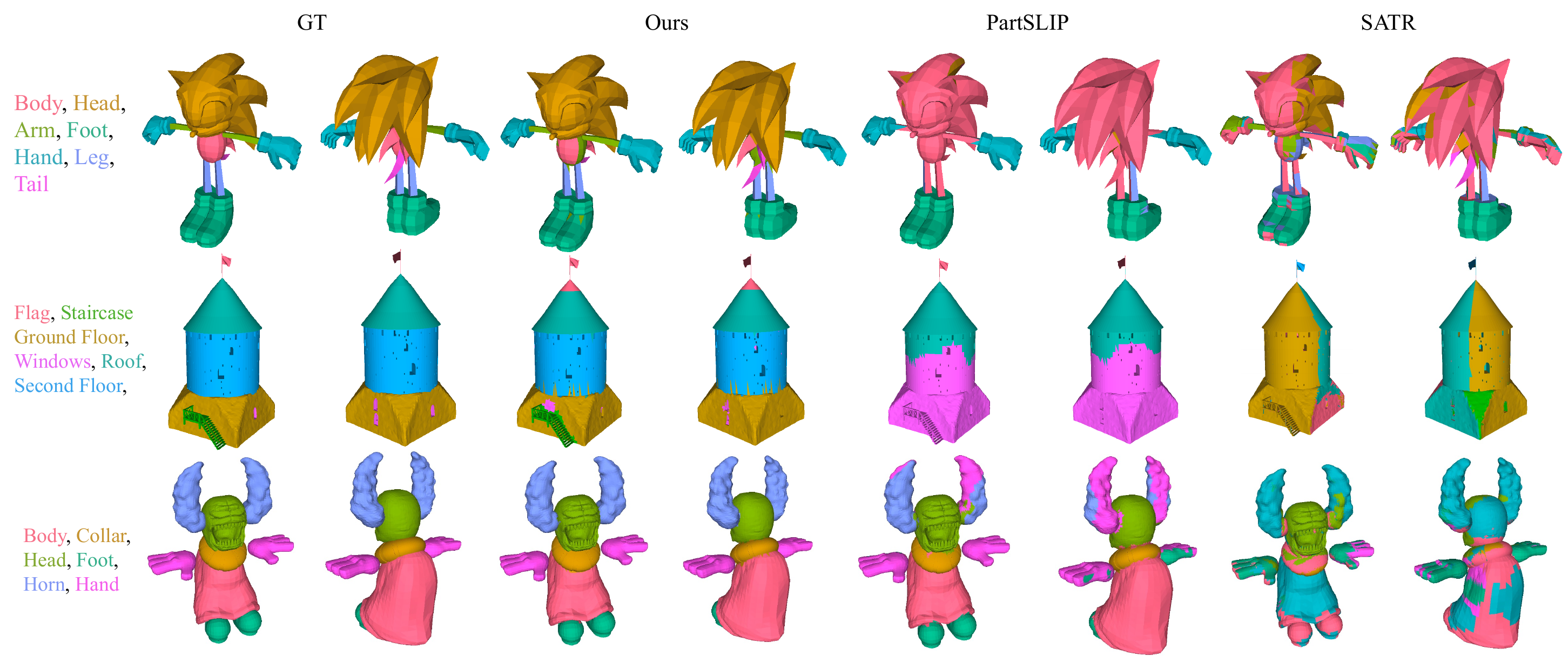}
\caption{Qualitative comparison with PartSLIP~\cite{liu2023partslip} and SATR~\cite{abdelreheem2023satr} in the semantic segmentation task on the PartObjaverse-Tiny dataset.}
\label{fig:exp_semseg}
\vspace{-5pt}
\end{figure*}

\noindent\textbf{Comparison with Existing Methods.} For the PartObjaverse-Tiny dataset, we evaluate our method against PointCLIP~\cite{zhang2022pointclip}, PointCLIPv2~\cite{zhu2023pointclipv2}, SATR~\cite{abdelreheem2023satr} and PartSLIP~\cite{liu2023partslip} for zero-shot semantic segmentation, as shown in \tabref{tab:sem_partobj}; against SAM3D~\cite{yang2023sam3d} and PartSLIP for zero-shot class-agnostic part segmentation in \tabref{tab:part_partobj}; and against PartSLIP for instance segmentation in \tabref{tab:ins_partobj}. For methods such as PartSlip and SATR, which utilize the GLIP detection model, the resulting segmentation often exhibits numerous blank areas. To address this, we employ the k-Nearest Neighbors (kNN) method, assigning the label of each face to that of the nearest face with predicted results. We present qualitative comparisons in \figref{fig:exp_semseg}. Note that our pre-training dataset excludes the 200 3D objects in PartObjaverse-Tiny for fair comparison.

We further compare our method with PointCLIPv2, PartSLIP, ZeroPS~\cite{xue2023zerops} and PartDistill~\cite{umam2023partdistill} on the PartNetE~\cite{liu2023partslip} dataset, as shown in \tabref{tab:sem_partnete}. We assign the label "others" for unlabeled areas in the PartNetE dataset.

\begin{table*}[!ht]
\centering

\resizebox{0.99\textwidth}{!}{

\begin{tabular}{c|c|cccccccc}
\hline
 Method  & Overall&  Human-Shape& Animals& Daily-Used& Buildings & Transportations & Plants& Food & Electronics         \\  \hline  
 
 PointCLIP     &5.4 &3.5 &4.5 & 6.5 &5.5 & 3.6 & 8.8 & 12.3 &5.6  \\

 PointCLIPv2   &9.5 &6.8 &10.0 &11.3 &8.4 &6.5 &15.8 &15.3 &9.9  \\
                      
 SATR          &12.3 &15.6 &16.5 &12.7 &7.9 &9.4 &17.2 &14.5 &9.7  \\ 

 PartSLIP     & 24.3& 39.3&  41.1& 19.0& 13.0& 17.1& 31.7& 17.3& 18.5  \\ 
  
 Ours  & \textbf{34.7}& \textbf{44.4}& \textbf{51.6}& \textbf{33.6}& \textbf{20.7}& \textbf{26.6}& \textbf{42.6}& \textbf{35.1}& \textbf{31.1}\\ \hline
\end{tabular}
}
\vspace{-5pt}
\caption{Zero-shot semantic segmentation on PartObjaverse-Tiny, reported in mIoU (\%). 
}
\vspace{-5pt}
\label{tab:sem_partobj}
\end{table*}
\begin{table*}[!ht]
\centering

\resizebox{\linewidth}{!}{

\begin{tabular}{c|c|cccccccc}
\hline
 Method  & Overall&  Human-Shape& Animals& Daily-Used& Buildings & Transportations & Plants& Food & Electronics         \\  \hline  

 PartSLIP      &35.2 &45.0 &50.1 &34.4 &22.5 &26.3 &44.6 &33.4 &32.0  \\ 

 SAM3D &43.6 &47.2 &45.0 &43.1 &38.6 &39.4 &51.1 &46.8 &43.8 \\
  
 Ours  & \textbf{53.7} & \textbf{54.4}& \textbf{59.0}& \textbf{52.1}& \textbf{46.2}& \textbf{50.3}& \textbf{60.7}& \textbf{59.8}& \textbf{54.5}\\ \hline
\end{tabular}
}
\vspace{-5pt}
\caption{Zero-shot class-agnostic part segmentation on PartObjaverse-Tiny, reported in mIoU (\%). 
}
\vspace{-5pt}
\label{tab:part_partobj}
\end{table*}
\begin{table*}[!ht]
\centering

\resizebox{0.99\textwidth}{!}{

\begin{tabular}{c|c|cccccccc}
\hline
 Method  & Overall&  Human-Shape& Animals& Daily-Used& Buildings & Transportations & Plants& Food & Electronics         \\  \hline 

 PartSLIP      &16.3 &23.0 &34.1 &13.1 &6.7 &10.4 &28.9 &7.2 &10.2 \\ 
  
 Ours & \textbf{30.2}& \textbf{36.9}& \textbf{43.7}& \textbf{29.0}& \textbf{19.0}& \textbf{21.4}& \textbf{38.5}& \textbf{39.4}& \textbf{27.7}\\ \hline
\end{tabular}
}
\vspace{-5pt}
\caption{Zero-shot instance segmentation on PartObjaverse-Tiny, reported in mAP50 (\%). 
}
\vspace{-5pt}
\label{tab:ins_partobj}
\end{table*}
\begin{table}[!ht]
\centering

\resizebox{0.47\textwidth}{!}{

\begin{tabular}{c|ccccc}
\hline
 Method  & PointCLIPv2 & PartSLIP & ZeroPS  & PartDistill &  Ours \\
 \hline 
   Overall & 16.1 &34.4&39.3&39.9&\textbf{41.2}  \\ \hline
\end{tabular}
}
\vspace{-5pt}
\caption{Zero-shot semantic segmentation on PartNetE~\cite{liu2023partslip}, reported in mIoU (\%). 
}
\vspace{-10pt}
\label{tab:sem_partnete}
\end{table}

\begin{figure*}
\centering
\includegraphics[width=0.99\linewidth]{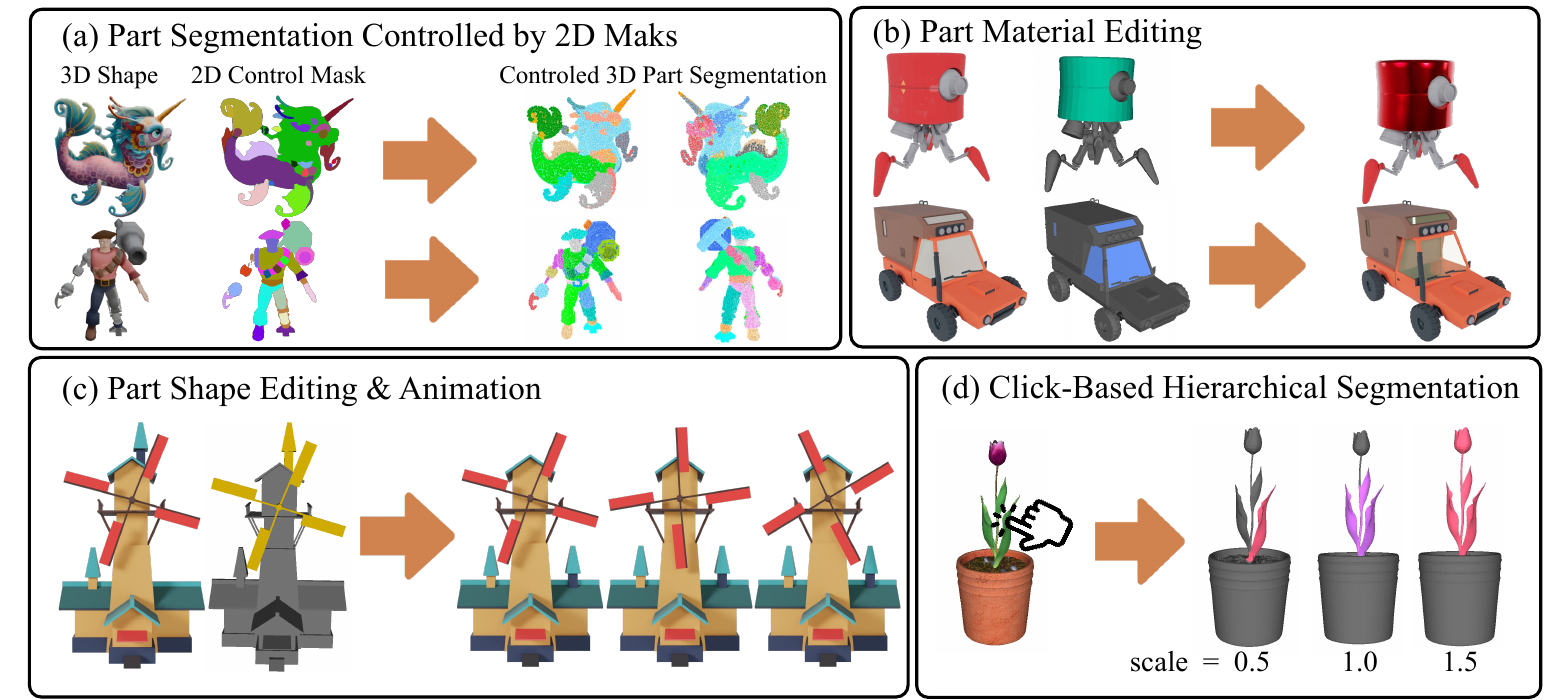}
\vspace{-7pt}
\caption{The resulting 3D part segmentation can directly support various applications, including part segmentation controlled by 2D masks, part material editing, part geometry editing, and click-based hierarchical segmentation.}
\vspace{-7pt}
\label{fig:part-editing}
\end{figure*}
\begin{table*}[!ht]
\centering

\resizebox{0.99\textwidth}{!}{%

\begin{tabular}{c|c|c|cccccccc}
\hline
 Method  & Pre-train Data& Overall&  Human-Shape& Animals& Daily-Used& Buildings & Transportations & Plants& Food & Electronics         \\  \hline 

 w.o. pre.  &-    &43.4 &48.5 &45.7 &44.9 &31.7 &37.2 &54.5 &48.1 &44.8  \\ 
 PTv3  &36k    &46.7 &50.9 &48.7 &47.8 &38.5 &43.0 &51.5 &52.0 &47.0  \\ 
 w.o. skip   &36k   &48.7 &51.1 &51.0 &49.0 &40.5 &44.3 &59.0 &53.1 &49.5  \\ 
 Ours  &36k    &50.5 &53.3 &53.4 &51.1 &41.6 &45.5 &58.7 &57.2 &51.8  \\ 
 Ours &200k & \textbf{53.7} & \textbf{54.4}& \textbf{59.0}& \textbf{52.1}& \textbf{46.2}& \textbf{50.3}& \textbf{60.7}& \textbf{59.8}& \textbf{54.5}\\ \hline
\end{tabular}
}
\vspace{-5pt}
\caption{Ablation study on PartObjaverse-Tiny, reported in mIoU (\%).}
\vspace{-10pt}
\label{tab:abl_partobj}
\end{table*}

\subsection{Ablation Analysis}
\begin{figure}
\centering
\includegraphics[width=0.9\linewidth]{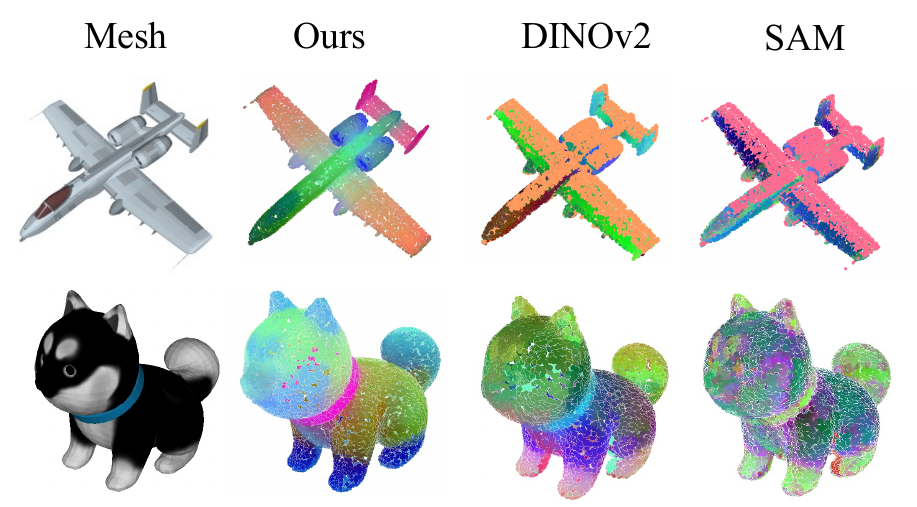}
\caption{Visualization and qualitative comparison of the features encoded by our backbone, DINOv2, and SAM. Due to the utilization of 3D information from point clouds, our backbone can produce more accurate and fine-grained visual semantic features.}
\vspace{-15pt}
\label{fig:encoded_emb}
\end{figure}

We conduct ablation studies on SAMPart3D, and the quantitative comparison are shown in \tabref{tab:abl_partobj}. To save pre-training time, we utilize a high-quality subset of Objaverse with 36k objects for ablation studies. We pre-train the original PTv3 backbone on this dataset. And we also pre-train our PTv3-object on this dataset with the same number of parameters as PTv3 for fair comparison.

\noindent\textbf{Necessity of Pre-training.} We ablate the proposed large-scale pretraining on Objaverse, distilling the knowledge from powerful 2D foundational model DINOv2. Without pre-training, we randomly initialize the PTv3-object backbone and retain other contents of the pipeline. Without pre-training, the model lacks rich semantic part information, which hinders its ability to effectively encode 3D objects. This limitation not only impacts the segmentation results but also leads to unstable training.

\noindent\textbf{PTv3-object v.s. PTv3.} We modify the original PTv3 backbone to PTv3-object, enhancing the model's encoding capabilities and ensuring the effective transmission of information at each point cloud. The second and fourth rows of \tabref{tab:abl_partobj} show comparison of PTv3-object and PTv3.

\noindent\textbf{Significance of Long Skip Connection.}
When training the grouping field, we freeze the backbone and only train MLPs for efficient training. At this stage, without the long skip connection, the model can only accept 3D embedding inputs rich in part semantic information from our backbone, but lacks direct input of 3D information. This leads to difficulty in training convergence, affecting final results.

\noindent\textbf{DINOv2's Feature v.s. SAM's Feature for Pre-training.} We utilize DINOv2 as our teacher model for pre-training, since DINOv2 contains extensive semantic information of the part and the whole object. We also attempt to directly utilize the features of SAM for pre-training, but find that the 3D backbone can not learn any knowledge that is helpful for part segmentation or perceiving objects. By pre-training on 3D large-scale data with more 3D spatial information inputs, our backbone's encoding capability exceeds the direct fusion of DINOv2's feature. We show qualitative comparison of our backbone's encoding feature, DINOv2's fusion feature and SAM's fusion feature in \figref{fig:encoded_emb}.

\subsection{Applications}

Recent several works~\cite{tertikas2023generating,yang2024dreamcomposer,zhang2024mapa,fang2024make,chung20243dstyleglip,qi2024tailor3d} attempt to generate or edit the style or material of parts for 3D objects. However, due to the lack of accurate 3D segmentation methods, most of these methods are designed to use 2D segmentation methods or traditional simple 3D segmentation methods instead. This limitation may cause these methods to struggle with complex objects or generate inconsistencies in 3D editing. Our model is capable of segmenting any 3D object at various scales, thereby serving as a robust tool for 3D generation and editing. At the same time, our method can serve as a pipeline for creating 3D part data, to create more 3D part assets for training 3D perception and generating models. 

\noindent\textbf{Part Segmentation Controlled by 2D Masks.} As illustrated in \figref{fig:part-editing} (a), our model seamlessly adapt to part segmentation guided by 2D segmentation masks. By utilizing 2D masks from a given viewpoint, we substitute SAM's masks with these input masks and compute a scale factor for the visible points within that view for inference.

\noindent\textbf{Part Material Editing.} Leveraging precise segmentation results for 3D objects, our approach enables the customization and editing of materials and styles for individual components. This greatly improves the adaptability of 3D models to different design needs, enabling detailed personalization and optimization of textures. As shown in \figref{fig:part-editing} (b), we can edit the material of parts for realistic outcomes.

\noindent\textbf{Part Shape Editing and Part Animation.} The 3D object part segmentation results produced by our model can be directly applied to part shape editing tasks. By using these segmentation results, we can modify selected components within Blender, streamlining the workflow for 3D model refinement and customization. As shown in \figref{fig:part-editing} (c), we can use the segmentation results of the windmill and chimney to alter the shape of object parts. Similarly, we can adjust the angle of the windmill to showcase part animation.

\noindent\textbf{Click-based Hierarchical Segmentation.} SAMPart3D is capable of accepting scale control to segment objects from coarse to fine. Therefore, similar to GraCo~\cite{zhao2024graco} in 2D, our model can accept a click in 3D space with a scale value to perform hierarchical segmentation of 3D objects. As shown in the \figref{fig:part-editing} (d), given a click at a location, the segmented area can be adjusted by controlling the scale.

\vspace{-1pt}
\section{Conclusion and Discussions}\label{sec:conclusion}
\vspace{-1pt}
We propose SAMPart3D, a zero-shot 3D part segmentation framework that can segment 3D objects into semantic parts at multiple granularities. 
Additionally, we introduce a new 3D part segmentation benchmark, PartObjaverse-Tiny, to address the shortcomings in diversity and complexity of existing annotated datasets. Experimental results demonstrate the effectiveness of SAMPart3D. 

{
    \small
    \bibliographystyle{ieeenat_fullname}
    \bibliography{main}
}

\clearpage
\section{Supplemental Material}

\subsection{Implementation Details}\label{subsec:implementation}
In the large-scale pre-training stage, we train the PTv3-object backbone on 200K high-quality objects from Objaverse with a batch size of 32, taking 7 days on eight A800 GPUs. Each object is rendered from 36 views, including 6 fixed views (+x, +y, +z, -x, -y, -z) and 30 random views. For each iteration, we pick 6 fixed views and 2 random views to cover most of the area for each object. We use the pre-trained DINOv2-ViTS14 model to encode the rendered images into visual features. These features are then upsampled by the FeatUp~\cite{fu2024featup} model to obtain pixel-wise features as supervision.

In the sample-specific fine-tuning stage, we initially sample 15K point clouds on the mesh surface for the object of interest. We render 36 views for the object, consistent with the settings used during the pre-training stage. We input the 2D rendered images into SAM to obtain the 2D segmentation masks. For each iteration, we randomly pick 90 rendered images (with replacement) and sample 256 valid pixels from each image, obtaining 23,040 3D points mapped from pixels as inputs.
The modules used for scale-conditioned grouping and long skip connection each employ 6-layer and 4-layer MLPs, with hidden dimensions set to 384. This stage requires 1 minute to generate masks using SAM, followed by 5 minutes of training MLPs.

After the fine-tuning stage, we can obtain the segmentation-aware features of 3D point cloud conditioned on a scale. We use the clustering algorithm HDBSCAN~\cite{mcinnes2017hdbscan} for feature grouping, and utilize GPT-4o~\cite{gpt4o} for per-part semantic querying.

\subsection{Ablation Analysis of Segmentation Scale}
Figure 4 in the paper presents a visualization of segmentation results across different scale factors. For the quantitative analysis of scale, we use five scale values [0.0, 0.5, 1.0, 1.5, 2.0] for each object, automatically selecting the result closest to the ground truth for evaluation. We conduct measurements using five values: [0.0, 0.5, 1.0, 1.5, 2.0] and compare them with our mixed scale results. Since the dataset is manually annotated, the number of suitable scales for different objects varies, leading to differences between individual and mixed scale results. The results of class-agnostic part segmentation results at different scales is shown in \tabref{tab:abl_scale}.

\begin{table*}[!ht]
\centering

\resizebox{\linewidth}{!}{

\begin{tabular}{c|c|cccccccc}
\hline
 Method  & Overall&  Human-Shape& Animals& Daily-Used& Buildings & Transportations & Plants& Food & Electronics         \\  \hline  

 0.0 & 49.1 & 52.6 & 54.8 & 45.3 & 42.4 & 47.8 & 55.2 &42.5 & 49.5\\

 0.5 &48.9 & 51.1 & 53.1 & 47.1 & 41.3 & 46.9 & 58.0 & 50.9 & 48.0 \\

 1.0 & 39.6 & 35.2 & 43.0 & 42.0 & 37.4 & 34.1 & 41.9 & 50.4 & 43.4\\

 1.5 & 31.5 & 26.7 &31.0 & 34.3 & 20.9 & 24.9 & 34.1 & 52.3 & 35.6 \\

 2.0 & 24.4 & 21.5 & 24.6 & 30.6 & 22.5 & 15.4 & 30.3 & 48.4 & 24.9\\
  
 mixed-scale  & \textbf{53.7} & \textbf{54.4}& \textbf{59.0}& \textbf{52.1}& \textbf{46.2}& \textbf{50.3}& \textbf{60.7}& \textbf{59.8}& \textbf{54.5}\\ \hline
\end{tabular}
}
\caption{Zero-shot class-agnostic part segmentation on PartObjaverse-Tiny across different scale values, reported in mIoU (\%). 
}
\label{tab:abl_scale}
\end{table*}

\subsection{Limitations}\label{subsec:discussion}
The training for the grouping field stage uses masks from SAM segmented at different scales. If some of these masks are inaccurate, it can affect the final results. Also, training the grouping field for each object is still slow. A better solution might be to utilize our method as a pipeline to annotate a large number of part data, and then train a large 3D part segmentation model.

\subsection{Visualization on PartNetE Dataset}
\begin{figure*}[htbp]
\centering
\includegraphics[width=\linewidth]{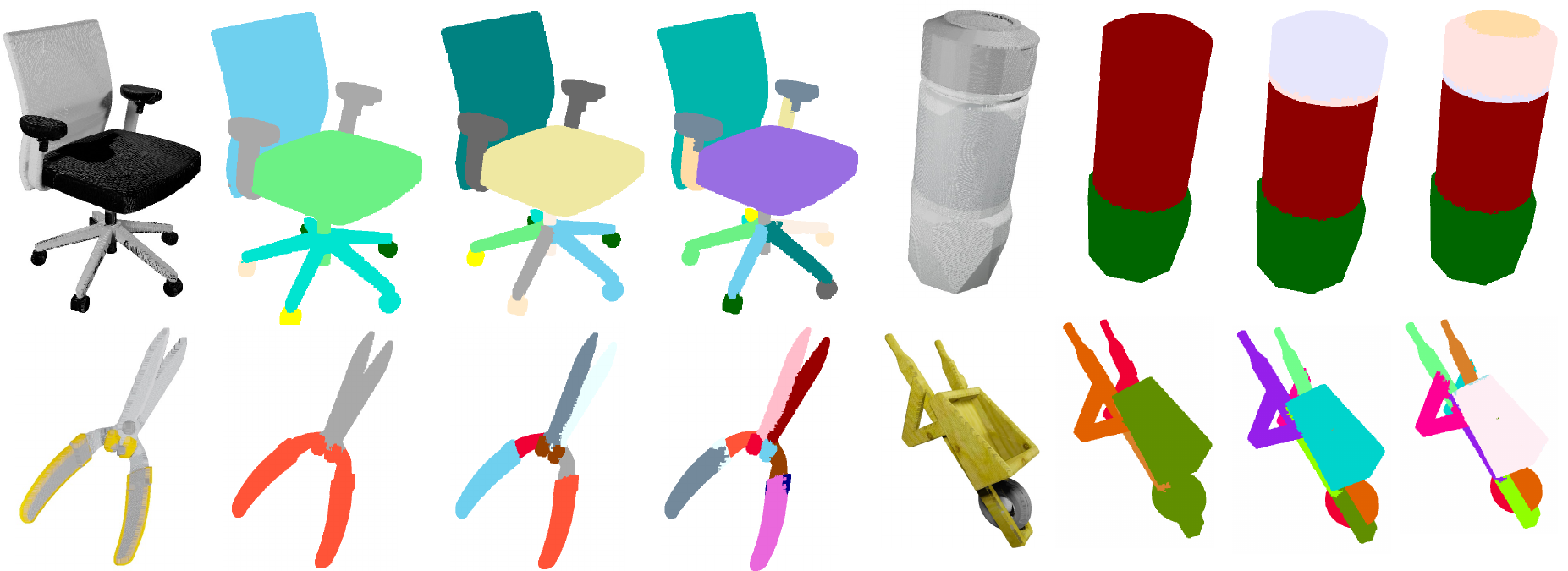}
\caption{Visualization of segmentation results on PartNetE dataset.}
\label{fig:supp_partnete}
\end{figure*}

We visualize the segmentation results of our SAMPart3D on the PartNetE~\cite{mo2019partnet} dataset in Fig.~\ref{fig:supp_partnete}. Compared to previous methods, our SAMPart3D can segment all fine-grained parts of 3D objects from PartNetE, even for parts that are not annotated in the dataset.

\subsection{More Qualitative Results of Our SAMPart3D}
\begin{figure*}
\centering
\includegraphics[width=\linewidth]{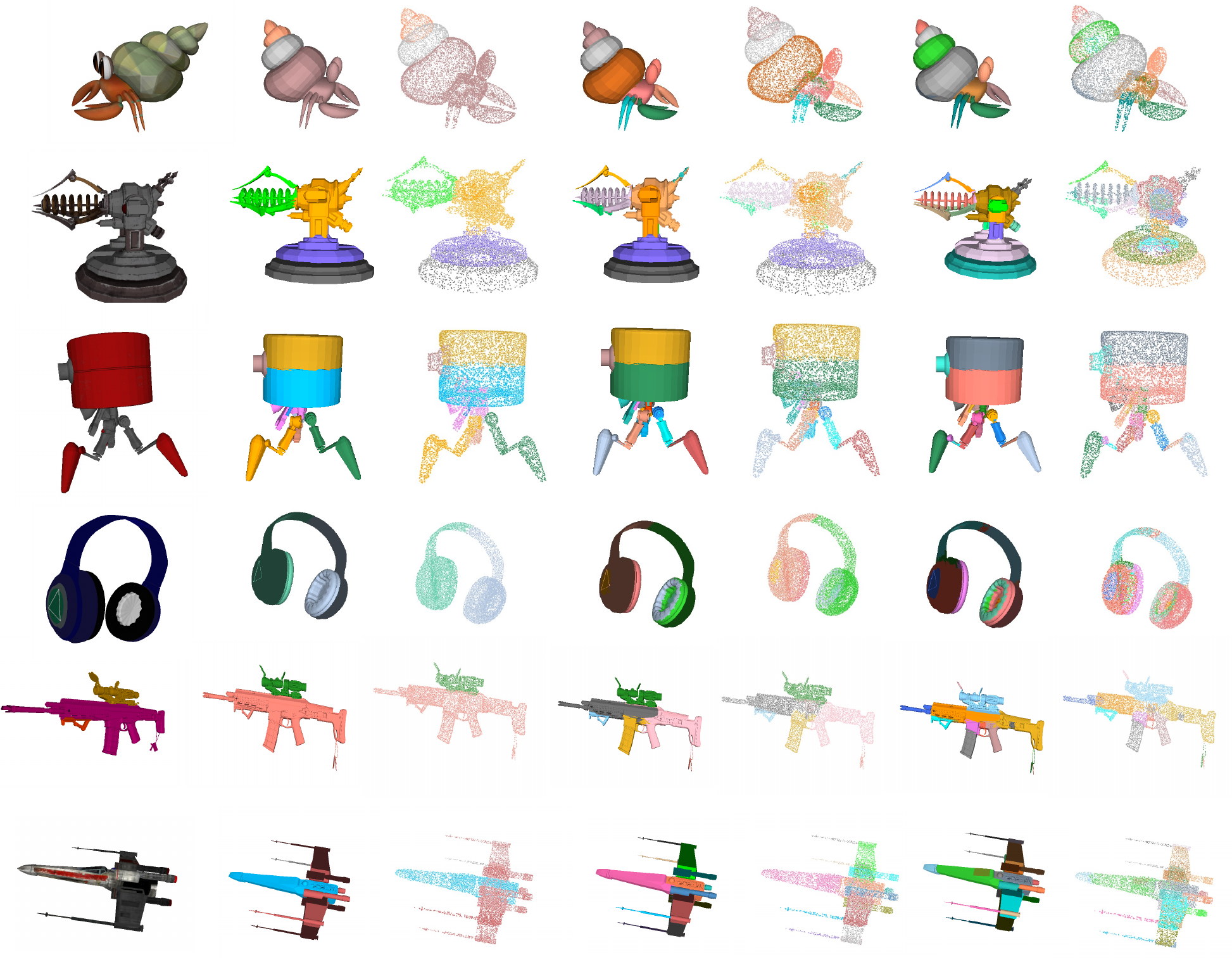}
\caption{Visualization of multi-granularity segmentation of point clouds and meshes.}
\label{fig:supp_res}
\end{figure*}

We show more visualization results of multi-granularity point clouds and meshes produced by our SAMPart3D in \figref{fig:supp_res}.

\subsection{More PartObjaverse-Tiny Visualization}
\begin{figure*}
\centering
\includegraphics[width=0.8\linewidth]{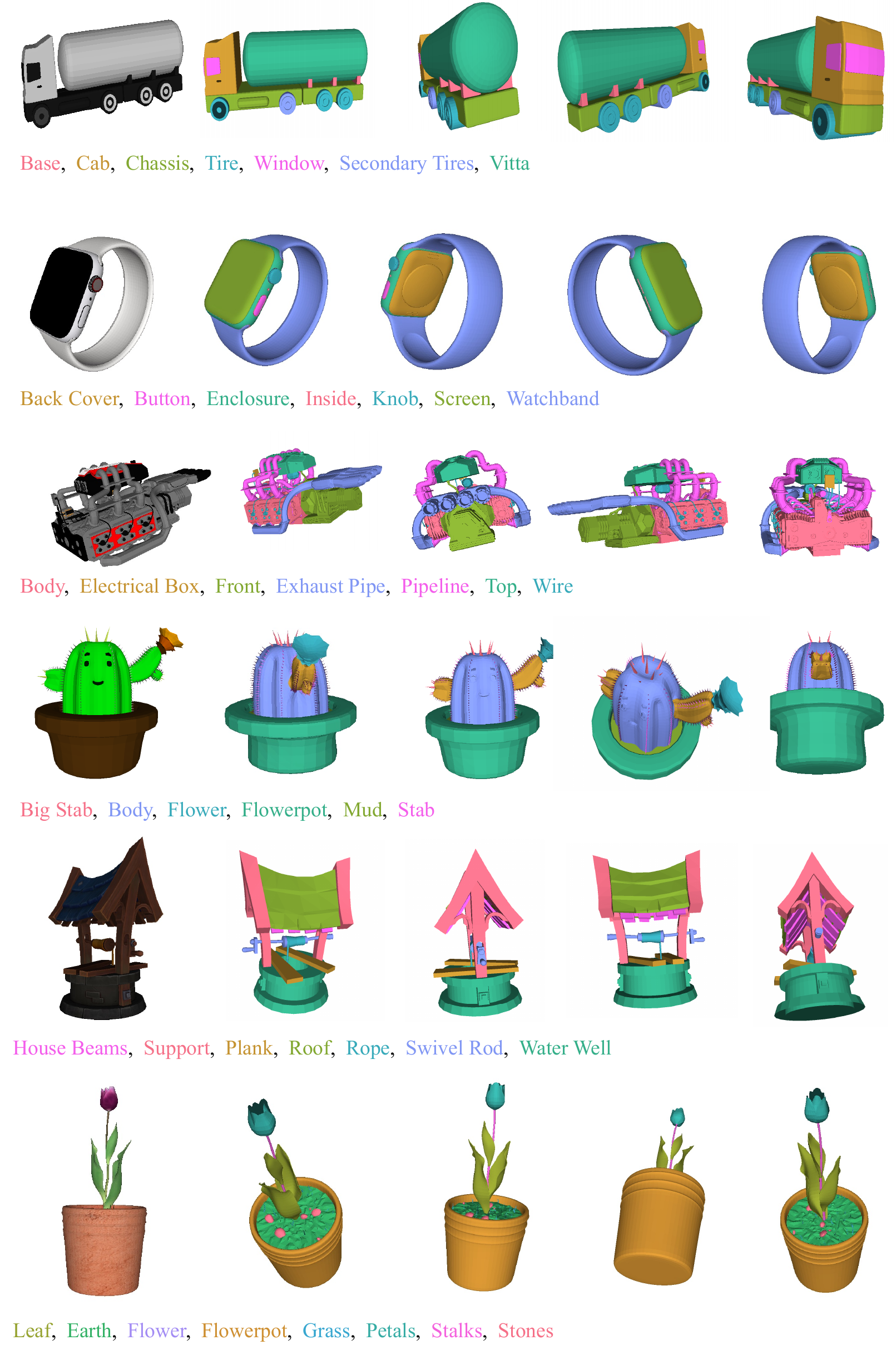}
\caption{Visualization of \textbf{PartObjaverse-Tiny} with part-level annotations with semantic labels for segmentation segmentation.}
\label{fig:supp_sem}
\end{figure*}

\begin{figure*}
\centering
\includegraphics[width=0.8\linewidth]{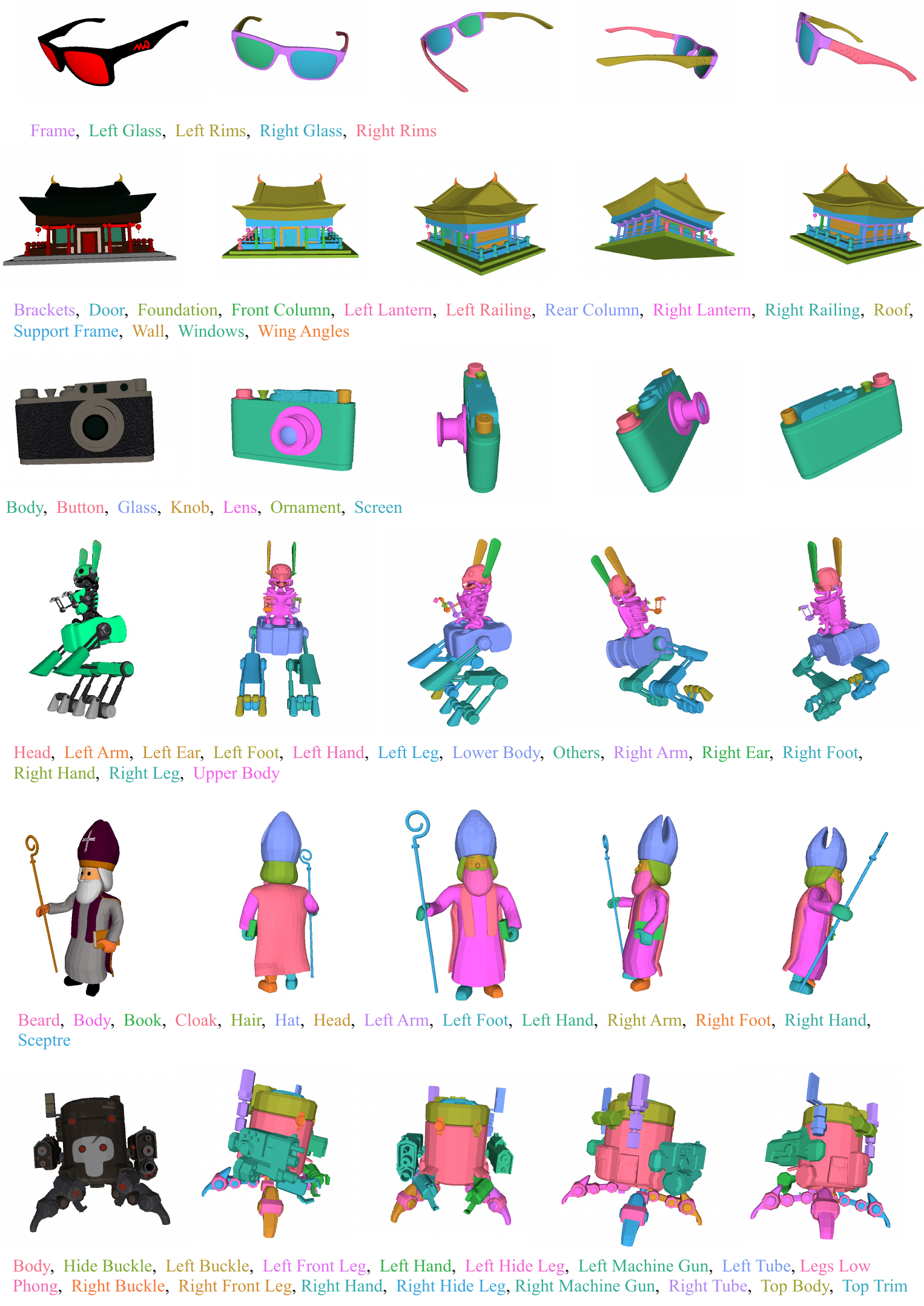}
\caption{Visualization of \textbf{PartObjaverse-Tiny} with part-level annotations with semantic labels for instance segmentation.}
\label{fig:supp_ins}
\end{figure*}

We demonstrate more visualization of dataset PartObjaverse-Tiny. We present more examples of semantic segmentation annotations in \figref{fig:supp_sem}, and instance segmentation annotations in \figref{fig:supp_ins}.

\clearpage

\clearpage

\clearpage

\end{document}